\documentclass[letterpaper, 10 pt, conference]{ieeeconf}  
\pdfoutput=1
\usepackage{times}
\usepackage{hyperref} 
\usepackage{amssymb}

\usepackage{amsfonts}
\usepackage{amsmath}
\usepackage{graphicx}
\usepackage{algorithm,algpseudocode}
\usepackage{xcolor}  
\usepackage{booktabs}  
\usepackage{array}     
\usepackage{cite}
\usepackage{float} 
\usepackage{subfigure}
\usepackage{subcaption}
\usepackage{caption}
\usepackage{balance} 
\usepackage{wrapfig}

\IEEEoverridecommandlockouts                              

\title{\LARGE \bf
Teacher Motion Priors: Enhancing Robot Locomotion over Challenging Terrain
}

\author{
    Fangcheng Jin$^{1, 2}$, Yuqi Wang$^{3}$, Peixin Ma$^{3}$, 
    Guodong Yang$^{1, 2}$, Pan Zhao$^{3}$, En Li$^{1, 2}$, Zhengtao Zhang$^{1, 2}$
    \thanks{*Corresponding Author: Guodong Yang}%
    \thanks{*This work was supported in part by the National Natural Science Foundation of China under Grant 62273344, 
                                and in part by Beijing Zhongke Huiling Robot Technology Co., LTD}
    \thanks{$^{1}$The authors are with the School of Artificial Intelligence, 
        University of Chinese Academy of Sciences, Beijing, Beijing 100000, China 
        {\tt\small jinfangcheng23@mails.ucas.ac.cn}}%
    \thanks{$^{2}$The authors are with Institute of Automation, Chinese Academy of Sciences, 
            Beijing, Beijing 100000, China 
        {\tt\small \{guodong.yang, en.li, zhengtao.zhang\}@ia.ac.cn}}%
    \thanks{$^{3}$The authors are with Beijing Zhongke Huiling Robot Technology Co., LTD, Beijing 100192, China}%
}

\begin{document}

\maketitle
\thispagestyle{empty}
\pagestyle{empty}

\begin{abstract}

    Achieving robust locomotion on complex terrains remains a challenge due to high-dimensional control and environmental uncertainties. 
    This paper introduces a teacher-prior framework based on the teacher-student paradigm, integrating imitation and auxiliary task learning to improve learning efficiency and generalization. 
    Unlike traditional paradigms that strongly rely on encoder-based state embeddings, our framework decouples the network design, simplifying the policy network and deployment. 
    A high-performance teacher policy is first trained using privileged information to acquire generalizable motion skills. 
    The teacher’s motion distribution is transferred to the student policy, which relies only on noisy proprioceptive data, via a generative adversarial mechanism to mitigate performance degradation caused by distributional shifts. 
    Additionally, auxiliary task learning enhances the student policy’s feature representation, speeding up convergence and improving adaptability to varying terrains. 
    The framework is validated on a humanoid robot, showing a great improvement in locomotion stability on dynamic terrains and significant reductions in development costs. 
    This work provides a practical solution for deploying robust locomotion strategies in humanoid robots.
\end{abstract}


\section{Introduction}

Robust locomotion on complex terrains remains a core challenge in robotics due to high-dimensional control and environmental uncertainties. 
Early model-based control methods enabled basic walking on challenging terrains~\cite{ref1, ref2, ref3, ref4, ref5} and were extended to humanoid robots for various tasks~\cite{ref6, ref7, ref8}, 
but these approaches often lack adaptability in real-world scenarios. 
Recent advancements in reinforcement learning (RL) have shown promise for addressing complex control problems~\cite{ref9, ref10, ref11, ref16}, 
though applying RL to humanoid robots remains difficult due to their high degrees of freedom and the need for robust performance on dynamic terrains. 
The \textit{teacher-student paradigm} has emerged as a solution, where a high-performance teacher policy is trained using privileged information and transferred to a student policy that relies on proprioceptive inputs~\cite{ref12, ref17, ref18, ref19, ref26}. 
This approach enables efficient sim-to-real deployment, but still faces challenges such as distributional shift and network complexity.

Several improvements have been proposed, including \textit{Regularized Online Adaptation (ROA)} and \textit{Collaborative Training of Teacher-Student Policies (CTS)}\cite{ref20, ref22}, but these methods still struggle with distributional shift and network structure dependency, limiting their generalization ability. 
To address these issues, \textit{Generative Adversarial Imitation Learning (GAIL)}\cite{ref23} leverage adversarial training to alleviate distributional shift and decouple the student policy from the teacher’s network. 
Extensions like \textit{Adversarial Motion Priors (AMP)} further enhance motion generation by evaluating state transitions~\cite{ref21}, allowing the control strategy to generate stylized movements.
Additionally, \textit{Multi-Task Learning (MTL)}\cite{ref24, ref25} has been integrated into RL to accelerate training and improve generalization by enhancing feature representations~\cite{ref34, ref35, ref36}.

In this work, we propose a novel teacher-student framework, \textit{Teacher Motion Priors (TMP)}, that integrates generative adversarial mechanisms and auxiliary task learning to tackle distributional shift, network dependency, and limited generalization. Our key contributions include:
\begin{itemize}
\item \textbf{High-performance teacher policy}: We train a robust teacher policy with privileged information and large-scale networks to enable generalizable locomotion in complex environments.
\item \textbf{Generative adversarial knowledge transfer}: We transfer the teacher’s behavior distribution to the student policy, mitigating distributional shift and decoupling network structures.
\item \textbf{Auxiliary task learning for student policy}: We enhance feature representation, accelerate training, and improve generalization across dynamic terrains.
\item \textbf{Real-world validation}: The trained student policy is deployed on a full-scale humanoid robot, showing significant improvements in locomotion stability and robustness on dynamic terrains.
\end{itemize}

Our experiments on a humanoid robot platform demonstrate superior learning performance, enhanced tracking accuracy, and reduced Cost of Transport (CoT) compared to mainstream methods. 
The following sections present our method and experimental results in detail.

\section{Teacher Motion Priors}

\begin{figure*}[ht]
    \centering
        \includegraphics[width=0.9\textwidth]{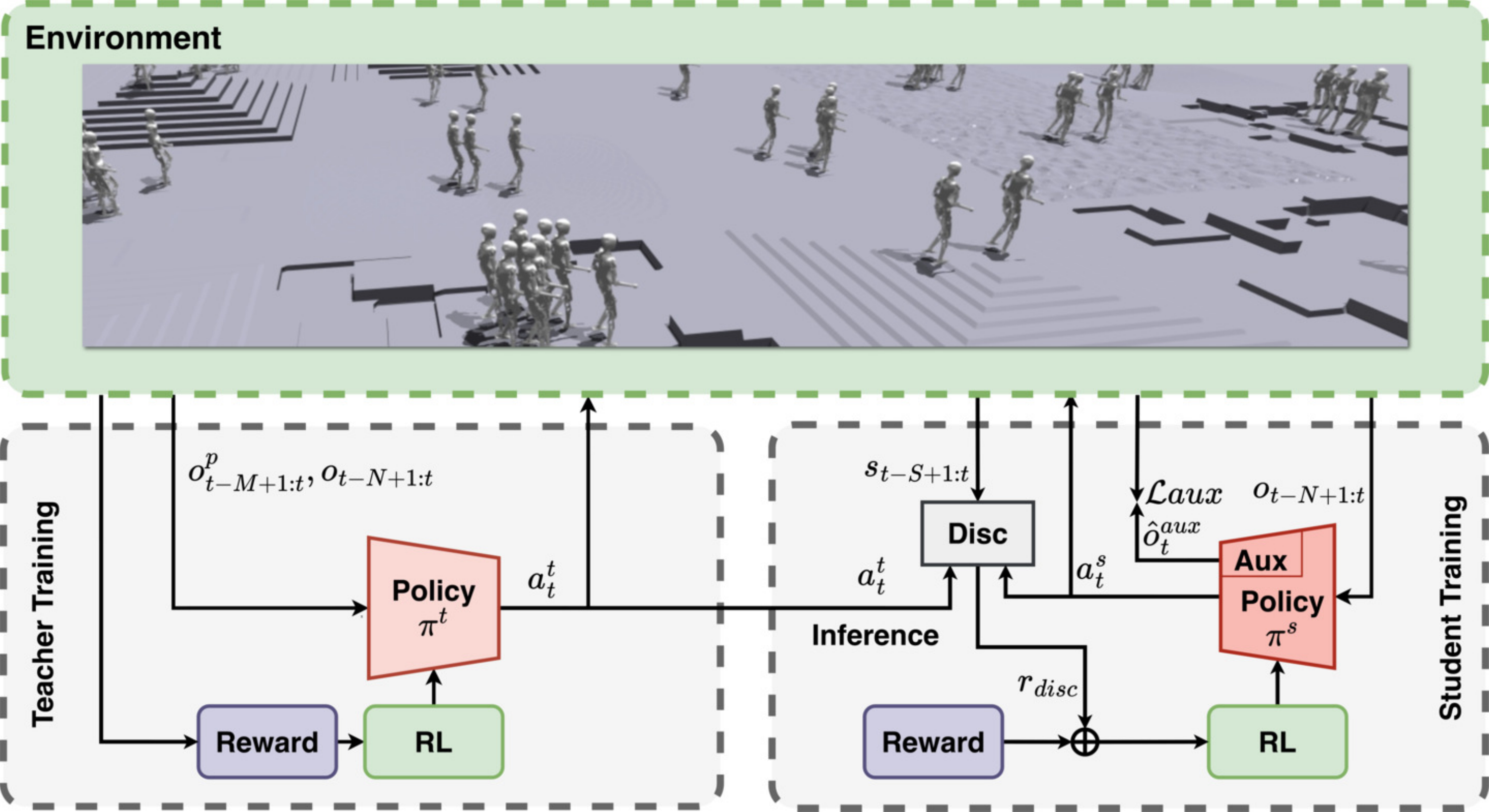} 
    \caption{TMP Training Process}
    \label{Fig.2}
 \end{figure*}

The training of the TMP framework consists of two stages.
As illustrated in Fig.~\ref{Fig.2}, the teacher phase on the bottom left is performed first, followed by the student phase on the bottom right.
In this section, we first present the problem formulation, followed by the proposed algorithmic framework.

\subsection{Humanoid Locomotion and Reinforcement Learning}

Our approach models the humanoid locomotion problem as a partially observable Markov decision process (POMDP), defined by the tuple $ \langle \mathcal{S, A, O, T, R, \gamma} \rangle $. 
Here,  $\mathcal{S}$  is the state space,  $\mathcal{A}$  is the action space,  $\mathcal{T}(s{\prime} | s, a)$  is the state transition function,  $\mathcal{R}(s, a, s{\prime})$  is the reward function, and  $\mathcal{O}$  is the observation space, representing partial environmental information. 
The discount factor  $\mathcal{\gamma} \in [0, 1]$  balances immediate and future rewards.

In simulated environments, the agent has full access to the state space, but in real-world scenarios, the agent only observes  $o \in \mathcal{O}$ , which may be incomplete or noisy. 
To address this, the policy $\pi(a | o_{\leq t})$  maps historical observations to actions, approximating the true state.

The objective is to find an optimal policy $\pi^*$ that maximizes the expected cumulative discounted reward:

\begin{equation}
J(\pi) = \mathbb{E}{\pi} \left[ \sum_{t=0}^{\infty} \mathcal{\gamma}^t \mathcal{R}(s_t, a_t, s_{t+1}) \right].
\end{equation}

Our framework employs proximal policy optimization (PPO) with an actor-critic architecture, replacing supervised learning with a generative adversarial approach for student policy training. 
This enables the student to mimic the teacher policy, achieving robust locomotion even without privileged information.

At time step $ t $, the proprioceptive observation $ o_t \in \mathbb{R}^n $ and privileged information $ o^p_t \in \mathbb{R}^m $ are combined into the full state $ s_t = [o_t, o^p_t] \in \mathbb{R}^{m+n} $. 
To enhance generalization, Gaussian noise is added to the proprioceptive observation input of the actor at both stages, while the privileged observation remains noise-free. 
The policy network outputs the action $ a_t \in \mathbb{R}^i $, where $ i $ is the number of controllable joints. 
The action controls the joint positions by being processed through a PD controller.
Superscripts $ (\cdot)^t $ and $ (\cdot)^s $ distinguish between teacher and student components, respectively.

\subsection{Teacher Policy}

In the teacher policy training, both privileged information and proprioceptive data are input into the teacher policy to guide robust locomotion strategy learning. 
To improve learning, we use frame stacking, where the teacher policy  $\pi^t$  takes $ N $ frames of proprioceptive data $ o_{t-N+1:t} \in \mathbb{R}^{N \times n} $ and $ M $ frames of privileged information $ o^p_{t-M+1:t} \in \mathbb{R}^{M \times m} $.

The teacher policy employs an actor-critic architecture. 
The actor generates actions by receiving privileged information $ o^p_{t-M+1:t} $ and proprioceptive observations $ o_{t-N+1:t} $.
The critic receives $ M $ frames of noise-free state data $ s_{t-M+1:t} \in \mathbb{R}^{M \times (m+n)} $. 
Detailed architecture is shown in Table~\ref{tab:teacher_student_policy}.

Training follows the process outlined in Algorithm~\ref{alg:teacher_training}, where policy parameters are updated using gradient descent to minimize the loss function.

\begin{algorithm}[ht]
\caption{Teacher Training Process}
\label{alg:teacher_training}
\begin{algorithmic}[1]
\State Initialize environment and networks.
\For{$k = 0, 1, \dots$}
    \State Collect a set of trajectories using the latest policy.
    \State Compute the target returns \( \hat{R}_t \) and advantages \( \hat{A}_t \) using GAE.
    \For{each epoch $i = 0, 1, \dots$}
        \State Update policy parameters using gradient descent:
        \[
        \theta^t \leftarrow \theta^t - \alpha \cdot \text{clip}\big( \nabla_{\theta^t} \mathcal{L}_{\text{teacher}}, -\text{max\_grad}, \text{max\_grad} \big)
        \]
    \EndFor
\EndFor
\end{algorithmic}
\end{algorithm}

\subsubsection*{Loss Function Definition}
The teacher policy optimizes the following loss function:

    \begin{equation}
        \begin{aligned}
            \mathcal{L}_{\text{teacher}} = \mathcal{L}_{\text{clip}} + \lambda_v \mathcal{L}_v - \lambda_e \mathcal{L}_e
        \end{aligned}
    \end{equation}
where:
\begin{itemize}
    \item \( \mathcal{L}_{\text{clip}} \) is the clipped surrogate loss that stabilizes updates:
    
    \begin{equation}
        \begin{aligned}
            \mathcal{L}_{\text{clip}} &= \mathbb{E}_t [  \min ( r_t(\theta^t) \hat{A}_t, \\
            &\text{clip}(r_t(\theta^t), 1 - \epsilon, 1 + \epsilon) \hat{A}_t ) ]
        \end{aligned}
    \end{equation}
    
    \item $ \mathcal{L}_v $ is the value function loss, measuring the mean squared error between predicted value $ V_{\theta^t}(s_t) $ and the target return $ \hat{R}_t $, computed with generalized advantage estimation (GAE):
    
    \begin{equation}
        \mathcal{L}_v = \mathbb{E}_t \left[ ( V_{\theta^t}(s_t) - \hat{R}_t )^2 \right]
    \end{equation}
    
    \item $ \mathcal{L}_e $ is the entropy loss, encouraging exploration by promoting diverse action distributions:
    
    \begin{equation}
        \mathcal{L}_e = \frac{1}{N} \sum_{t=1}^N \mathcal{H}(\pi_\theta^t(\cdot | s_t))
    \end{equation}
    where $ \mathcal{H}(\pi_\theta^t(\cdot | s_t)) $ is the entropy:
    \begin{equation}
        \mathcal{H}(\pi_\theta^t(\cdot | s_t)) = -\sum_{a \in \mathcal{A}} \pi_\theta^t(a | s_t) \log \pi_\theta^t(a | s_t)
    \end{equation}
\end{itemize}

Entropy measures the uncertainty of the policy’s action selection. 
By adjusting $ \lambda_v $ and $ \lambda_e $, these terms balance exploration and convergence, optimizing the teacher policy for stable and efficient locomotion.

\subsection{Student Policy}

During student policy training, the student actor receives only proprioceptive data $ o_{t-N+1:t} $, while the critic remains similar to the teacher’s. 
Inspired by GAIL, we replace traditional supervised learning with a generative adversarial approach to help the student mimic the teacher’s behavior.

While collecting trajectories using the student policy, we also record the teacher’s response actions $ a^t_t $ at each state visited by the student. 
The discriminator $ \mathcal{D} $ receives the tuple $ (s_{t-S+1:t}, a_t) $, where $ s_{t-S+1:t} $ represents the last $ S $ frames of state information, and outputs $ p_\mathcal{D} \in [0, 1] $, indicating the likelihood that $ a_t $ is the teacher’s action.

The discriminator’s loss function is defined as:

\begin{equation}
    \mathcal{L}_{\text{disc}} = \lambda_{\text{pred}} \mathcal{L}_{\text{pred}} + \lambda_{\text{grad}} \mathcal{L}_{\text{grad}} + \lambda_{\text{weight}} \mathcal{L}_{\text{weight}}
\end{equation}
where:
\begin{itemize}
    \item \textbf{Prediction Loss}: The binary cross-entropy (BCE) loss classifies whether the trajectory originates from the teacher or the student:

    \begin{equation}
        \begin{aligned}
            \mathcal{L}_{\text{pred}} &= - \mathbb{E}_{\tau_t \sim \pi_{\text{teacher}}} \left[ \log \mathcal{D}(\tau_t) \right] \\
            &\quad - \mathbb{E}_{\tau_s \sim \pi_{\text{student}}} \left[ \log(1 - \mathcal{D}(\tau_s)) \right]
        \end{aligned}
    \end{equation}

    where \( \tau_t = \langle (s_{t-S+1:t}, a^t_t) \rangle_{t=0}^{T} \) and \( \tau_s = \langle (s_{t-S+1:t}, a^s_t) \rangle_{t=0}^{T} \) are the teacher and student trajectories, respectively.
    
    \item \textbf{Gradient Regularization}: This term penalizes large gradients to avoid overfitting:

    \begin{equation}
        \mathcal{L}_{\text{grad}} = \lambda_{\text{grad}} \mathbb{E}_{\tau \sim \pi_{\text{teacher}} \cup \pi_{\text{student}}} \left[ \| \nabla_{\tau} \mathcal{D}(\tau) \|^2 \right]
    \end{equation}
    where \( \tau \) denotes trajectories sampled from both the teacher and student policies, and \( \lambda_{\text{grad}} \) is the regularization coefficient.

    \item \textbf{Weight Regularization}: An L2 penalty on the discriminator’s weights improves generalization:
    \begin{equation}
        \mathcal{L}_{\text{weight}} = \lambda_{\text{weight}} \| \theta_\mathcal{D} \|^2
    \end{equation}
    where \( \theta_\mathcal{D} \) are the discriminator parameters, and \( \lambda_{\text{weight}} \) controls the regularization strength.
\end{itemize}

To accelerate training and enhance feature representation in the earlier network layers, we incorporate auxiliary task learning. 
The auxiliary network $ \text{aux} $ shares the first $ N-2 $ layers with the actor network and predicts auxiliary observations $ \hat{o}^{\text{aux}}_t $. 
This shared structure enhances the student’s ability to learn noise distributions in proprioceptive inputs and guide feature extraction, improving performance. 

The auxiliary loss function \( \mathcal{L}_{\text{aux}} \) is defined as:
\begin{equation}
    \mathcal{L}_{\text{aux}} = \mathbb{E}_{t} \left[ \big\| \hat{o}^{\text{aux}}_t - o^{\text{aux}}_t \big\|_2^2 \right]
\end{equation}
where \( \hat{o}^{\text{aux}}_t \) is the predicted auxiliary observation and \( o^{\text{aux}}_t \) is the ground truth auxiliary observation.

The student policy network is denoted by $ \theta^s $. 
The training process is summarized in Algorithm~\ref{alg:student_training}.

\begin{algorithm}[ht]
\caption{Student Training Process}
\label{alg:student_training}
\begin{algorithmic}[1]
\State Initialize environment and networks.
\For{$k = 0, 1, \dots$}
    \State Collect trajectories \( \tau_{\text{student}} \) using the current student policy.
    \State Collect teacher trajectories \( \tau_{\text{teacher}} \) using \( \pi^t \).
    \State Compute student policy target returns \( \hat{R}_t \) and advantages \( \hat{A}_t \) using GAE.
    \For{each epoch $i = 0, 1, \dots$}
        \State Update student policy parameters using:
        \[
        \theta^s  \leftarrow \theta^s  - \alpha \cdot \text{clip}\big( \nabla_{\theta^s } \mathcal{L}_{\text{student}}, -\text{max\_grad}, \text{max\_grad} \big)
        \]
        \State Update discriminator parameters using:
        \[
        \theta_\mathcal{D} \leftarrow \theta_\mathcal{D} - \alpha \cdot \text{clip}\big( \nabla_{\theta_\mathcal{D}} \mathcal{L}_{\text{disc}}, -\text{max\_grad}, \text{max\_grad} \big)
        \]
    \EndFor
\EndFor
\end{algorithmic}
\end{algorithm}

The student policy optimizes the following loss function, which combines adversarial and auxiliary task losses:
\begin{equation}
    \mathcal{L}_{\text{student}} = \mathcal{L}_{\text{clip}} + \lambda_v \mathcal{L}_v - \lambda_e \mathcal{L}_e + \lambda_{\text{aux}} \mathcal{L}_{\text{aux}} + \lambda_{\text{disc}} \mathcal{L}_{\text{disc}}
\end{equation}
The terms \( \mathcal{L}_{\text{clip}} \), \( \mathcal{L}_v \), and \( \mathcal{L}_e \) follow the same definitions as in the teacher policy but are optimized with respect to $ \theta^s $. 
Here, $ \lambda_{\text{aux}} $ and $ \lambda_{\text{disc}} $ control the contributions of the auxiliary task and adversarial losses, respectively. 
During the deployment phase, only the student policy $\pi^s$ is utilized, without the auxiliary network or critic.

\subsection{Training Configuration}

We define the robot’s base and foot poses using a six-dimensional vector $ [x, y, z, \alpha, \beta, \gamma] $, where $ [x, y, z] $ is the position and $ [\alpha, \beta, \gamma] $ is the orientation in Euler angles. 
A gait cycle, $ C_T $, consists of two double support (DS) phases and two single support (SS) phases. 
The leg reference trajectory is generated using quintic polynomial interpolation for foot height \cite{ref28}. 
The phase mask $ I_p(t) $ indicates foot contact, with DS phases marked as [1, 1] and SS phases as [1, 0] or [0, 1].

The proprioceptive input $ o_t \in \mathbb{R}^{47} $ includes standard sensory data, a phase clock signal $(\sin(t), \cos(t))$, and command input $\dot{P}_{x y \gamma}$. 
The privileged information $ o^p_t \in \mathbb{R}^{213} $ comprises data not accessible during deployment, 
such as a 187-dimensional height map representing the distance from the terrain to the robot’s base over a $1.6 \, \text{m} \times 1.0 \, \text{m}$ area and feet contact detection $I_d(t)$. 
Auxiliary information, used exclusively during student training, consists of partial states data for auxiliary task learning. 
A single frame of observations is elaborated in Table~\ref{tab:observations}.

During teacher policy training, 3 frames of privileged information (213 dimensions each) and 15 frames of proprioceptive data (47 dimensions each) are concatenated and fed into the actor.
Simultaneously, 3 frames of state data (260 dimensions each) are input into the critic. 
For the student policy, the actor input consists of 15 frames of proprioceptive data, with the critic structure unchanged from the teacher’s. 
The discriminator uses 10 frames of state data and 12 for the action dimension. 
Detailed network architecture is provided in Table~\ref{tab:teacher_student_policy}.

To ensure robust locomotion, we use a game-inspired curriculum learning, as described in \cite{ref37} across four terrain types: 
slopes, rough, stairs, and discrete obstacles. 
Slopes range from $0^\circ$ to $22.92^\circ$, with rough slopes adding uniform noise (-5 to 5 cm) to simulate uneven surfaces. 
Stairs vary from 5 cm to 24.95 cm, and obstacles range from 5 cm to 24 cm. 
The curriculum progresses through difficulty levels from 0 to 20, with each level comprising 20 terrain instances to ensure balanced exposure. 
Each level includes 4 rough terrains, 4 discrete obstacles, 3 upslopes, 3 downslopes, 3 stair ascents, and 3 stair descents.
Robots start at level 0 and progress to more challenging conditions as they successfully complete each level.

During training, velocity commands are uniformly sampled within \([-1.5, 1.5]\) m/s. 
Once robots perform well on challenging terrains and maintain accurate velocity tracking, the velocity range is gradually increased to promote more agile locomotion.

The student policy includes an auxiliary task network that shares the first layer with the actor network. 
The actor outputs actions $ a_t \in \mathbb{R}^{12} $, controlling the legs. 
The auxiliary network predicts auxiliary observations $ o_t^{\text{aux}} \in \mathbb{R}^{48} $. 
The discriminator distinguishes between teacher and student trajectories using inputs $ \langle o^p_t, a_t \rangle $.

\subsection{Reward Design}

We design a unified reward system to promote stable, energy-efficient locomotion while following gait patterns and velocity commands. 
The reward components include: (1) tracking reward, (2) periodic gait reward, (3) foot trajectory reward, and (4) regularization terms.
Additionally, to distinguish whether an action originates from the student or the teacher, a discriminator reward is introduced during student training. 

The \textbf{tracking reward} encourages accurate execution of velocity commands $ \text{CMD}_{xyz} $ and $ \text{CMD}_{\alpha \beta \gamma} $ by penalizing velocity errors:

\begin{equation}
    \phi(e, w) = \exp(-w \|e\|^2)  
\end{equation}

where $ e $ is the velocity error and $ w $ controls the penalty magnitude.

The \textbf{periodic gait reward} enhances coordination by penalizing deviations from the expected foot contact pattern, ensuring alignment with the gait phase through the binary phase mask.

The \textbf{foot trajectory reward} maintains desired foot height during the swing phase to ensure obstacle clearance:

\begin{equation}
    r_{\text{fc}} = \sum_{\text{swing}} \left| h_{\text{feet}} - h_{\text{target}} \right|  
\end{equation}

where \( h_{\text{feet}} \) and \( h_{\text{target}} \) represent the actual and target foot heights, respectively.

The \textbf{regularization terms} penalize undesired behaviors, including large joint torques, high accelerations, and excessive foot contact forces. 
The collision penalty is:

\begin{equation}
    n_{\text{collision}} = \sum_i \mathbb{I}(F_i > F_{\text{threshold}})  
\end{equation}

where $ F_i $ is the contact force and $ F_{\text{threshold}} = 0.1\text{N} $. The indicator function $ \mathbb{I}(\cdot) $ returns $1$ if the condition is met. 

The discriminator reward $r_{disc}$ is derived from a probability distribution $p_{disc}$, which encourages the student to mimic the teacher’s policy as closely as possible:
\begin{equation}
    p_{disc}=softplus(-\mathcal{D}(s_t))
\end{equation}
A higher $p_{disc}$ value (closer to 1) indicates that the student’s actions resemble those of the teacher to a greater extent.
Detailed reward configuration is in Table~\ref{tab:rewards}.

\subsection{Domain Randomization}

To address the sim-to-real gap, we apply domain randomization during training by varying key environmental and robot parameters, such as ground friction, stiffness, payload, joint friction, and PD controller settings. 
These randomizations improve the policy’s generalization ability by simulating diverse deployment scenarios. 
For a full list of randomization parameters, refer to Table~\ref{tab:domain_randomization}.

\begin{figure*}[ht]
    \centering
        \includegraphics[width=0.99\textwidth]{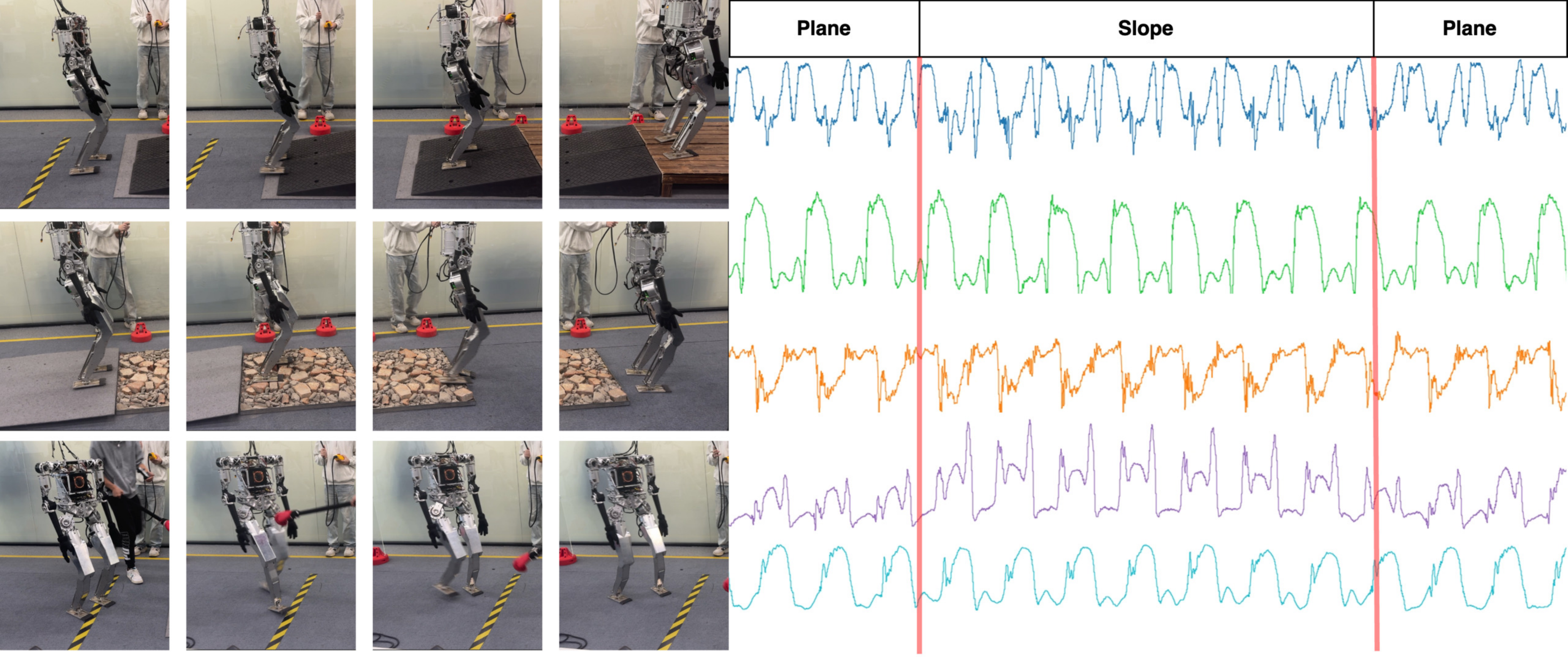} 
    \caption{CASBOT SE Multi-Terrain Testing in Real-World Environments. The first row shows slope testing, the second row presents trials on a brick-paved surface, and the third row demonstrates disturbance rejection.}
    \label{Fig.real}  
    \vspace{-12pt}
 
\end{figure*}

\section{Experiments}
\label{sec:experiments}
\subsection{Robot Platform Design} 

\begin{wrapfigure}[18]{o}{0.15\textwidth}
    \centering
        \includegraphics[width=0.9\linewidth]{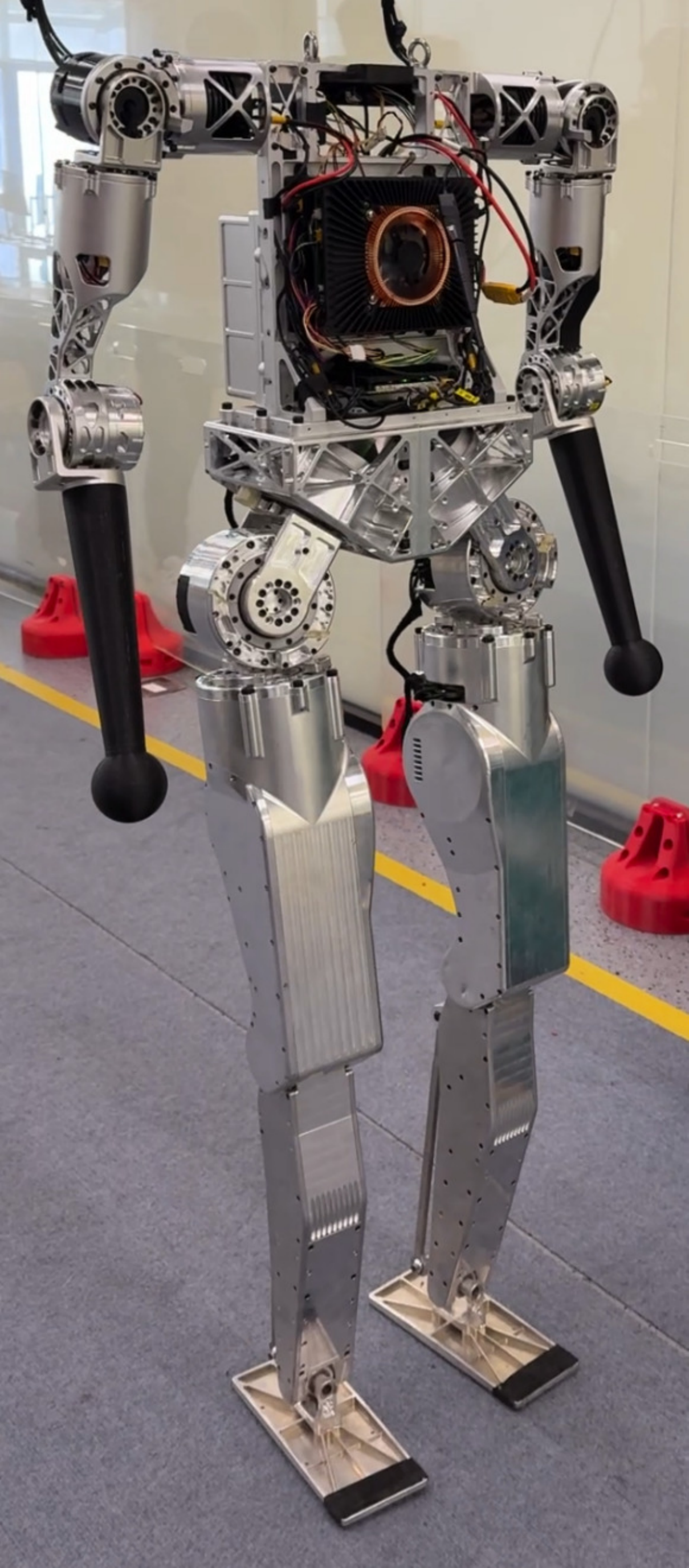} 
    \caption{Illustration of CASBOT SE.}
    \label{Fig.robot}   
\end{wrapfigure}

Our study is conducted on the CASBOT SE humanoid robot, developed by Beijing Zhongke Huiling Robot Technology Co., Ltd., as illustrated in Fig.~\ref{Fig.robot}. 
This full-sized platform stands 1.65 m tall and weighs 46 kg with 18 DoFs(6 on each leg, 3 on each arm). 
In this study, the arm joints are not utilized, and thus only 12 joints are controlled.
To achieve stable locomotion, a periodic reference trajectory for the feet is generated and solved using inverse kinematics to derive joint trajectories. 
A closed kinematic chain ankle mechanism, providing 2 DoFs, enhances robustness by reducing the impact of terrain irregularities on foot posture, improving stability on rough terrain.

\subsection{Evaluation Results}

We compared the performance, used \textit{Isaac Gym}, of several methods on the CASBOT SE as follows:
\begin{itemize}
\item \textbf{Oracle}: Policy trained with PPO, receiving  $s_{t-N+1:t}$  as input.
\item \textbf{Baseline}: PPO-trained policy with the actor receiving proprioceptive observations and the critic receiving privileged observations \cite{ref16}.
\item \textbf{Original teacher-student framework}: The teacher receives proprioceptive observations and latent code, and the student is trained using latent reconstruction and action imitation loss \cite{ref12}.
\item \textbf{Regularized Online Adaptation (ROA)}: Policy trained by integrating latent code between privileged and proprioceptive encoders \cite{ref20}.
\end{itemize}

All methods were trained in the actor-critic framework with identical configurations, network scale, and random seeds, evaluated over 10000 iterations. 
For the original teacher-student framework, 3000 iterations were allocated for the teacher, as in TMP. 
For its unique configurations, ROA was trained using the setup from \cite{ref20}.

\subsubsection*{Terrain Level Convergence}
\begin{figure}[thpb]
    \centering
        \includegraphics[width=3in]{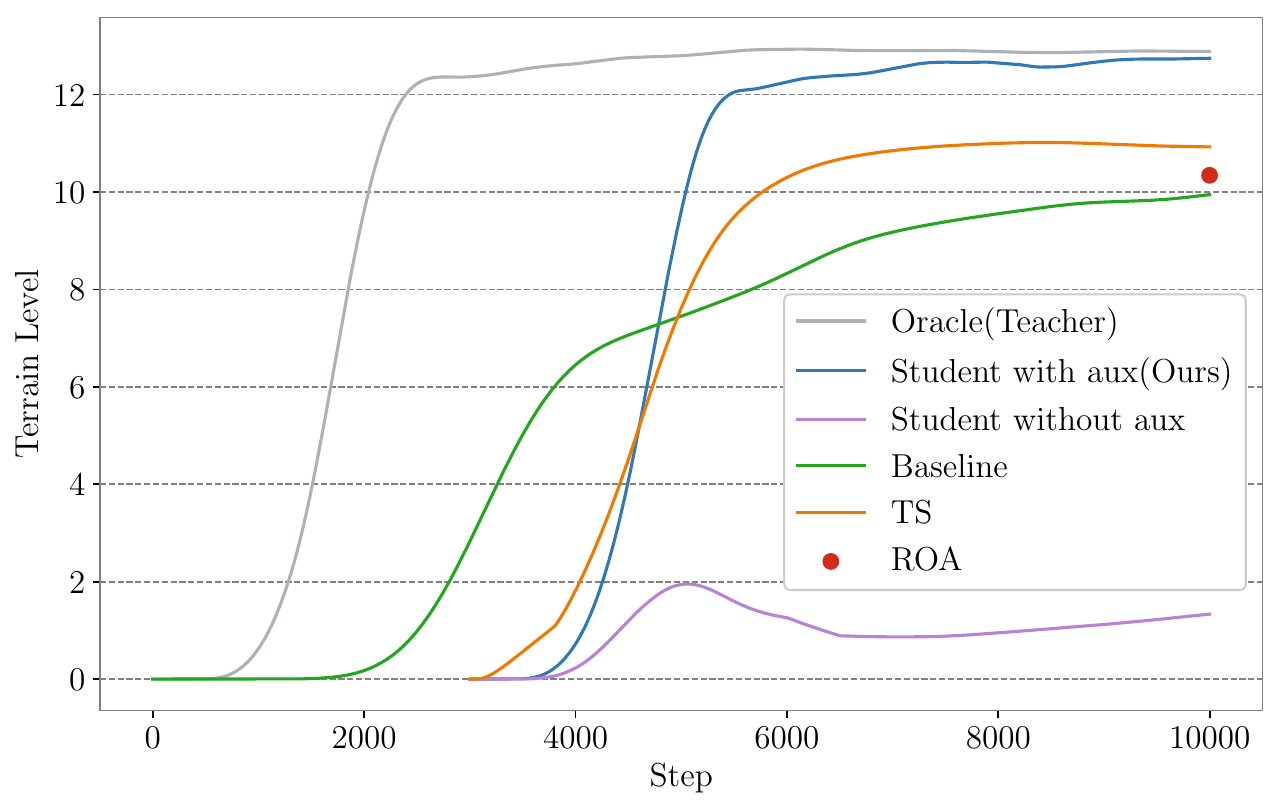} 
    \caption{Learning Curves of average terrain level}
    \label{Fig.4}
    \vspace{-10pt}
\end{figure}

We compared the performance of these methods in terms of terrain level, as shown in Fig.~\ref{Fig.4}. 
The curves, averaged over 10 seeds, represent the average terrain level of all agents at each training step, with the shaded area indicating the standard deviation.
With generative adversarial training and auxiliary task learning, the student policy closely matches the teacher policy. 
In contrast, the student policy without auxiliary task learning takes longer to converge. 
ROA’s terrain level curve does not effectively capture traversability due to its policy switching, but it shows a slight performance improvement over the baseline after 5000 iterations.
The final learning performance of TMP improves by 26.39\% and 17.20\% compared to TS and ROA.
We believe that improving the teacher policy, particularly enhancing the network architecture, can further benefit the student policy through TMP.

\subsubsection*{Tracking Accuracy}
\label{Tracking Accuracy}
\begin{figure}[thpb]
    \centering
        \includegraphics[width=3in]{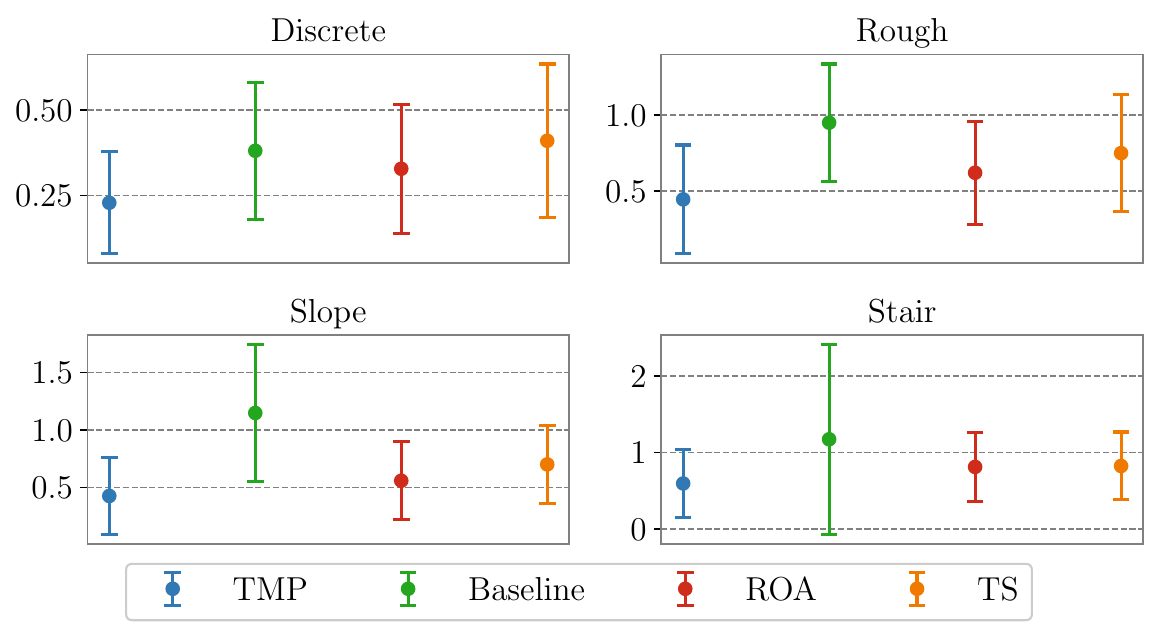} 
    \caption{Evaluation of average tracking error in 4 types of terrains.}
    \label{Fig.5}
    \vspace{-15pt}
\end{figure}

We evaluated velocity tracking across diverse terrains using 10240 uniformly distributed robots. 
Linear velocity commands were sampled from [-1.5, 1.5] m/s, and tracking errors were computed as  $|| \text{CMD}_{xy} - v_{xy} ||_2$.
Fig.~\ref{Fig.5} presents the average tracking performance, with tracking errors across 4 terrain types shown on the y-axis. 
Each subplot compares linear velocity tracking errors over 10 seeds.
TMP outperforms TS and ROA, reducing errors by 44.16\% and 30.25\% on discrete obstacles, 40.53\% and 28.16\% on rough slopes, 39.17\% and 23.71\% on slopes, 27.74\% and 26.66\% on stairs. 
While ROA achieves comparable terrain level performance to the baseline, it exhibits higher tracking accuracy.

\subsubsection*{CoT}
\begin{figure}[thpb]
    \centering
        \includegraphics[width=3in]{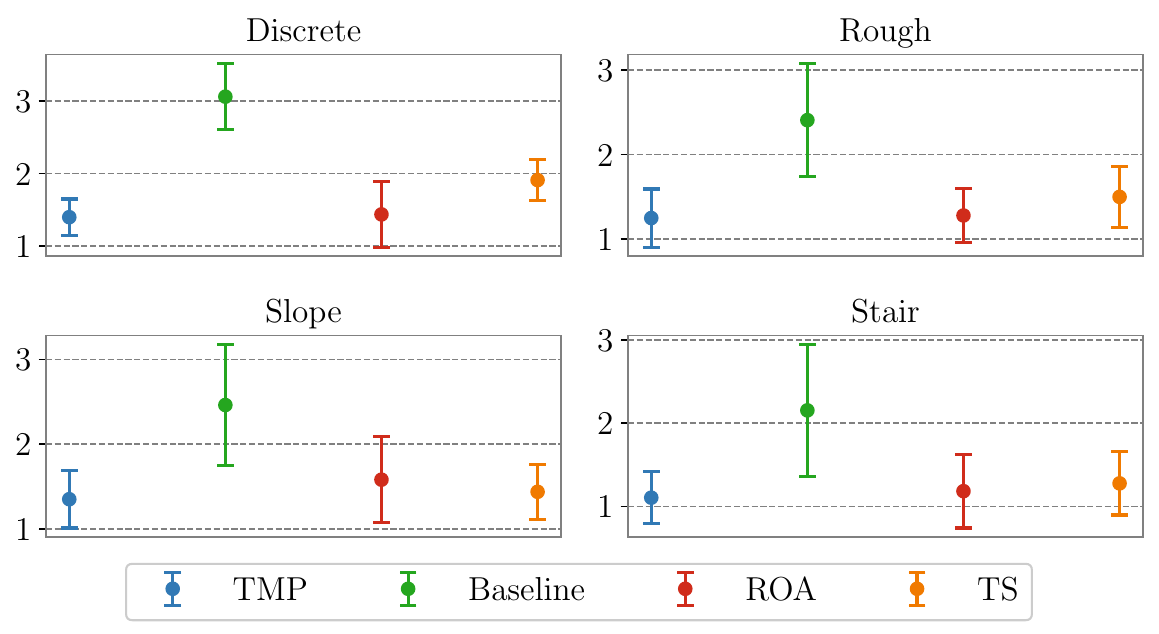} 
    \caption{Evaluation of average tracking error in 4 types of terrains.}
    \label{Fig.6}
    \vspace{-15pt}

\end{figure}

We evaluate the policy’s Cost of Transport (CoT), defined as ~\cite{ref12}, which quantifies the energy efficiency of the policy in controlling the robot.
We evaluate each policy using the same speed sampling and environmental setup as described in the ~\ref{Tracking Accuracy} section.
Fig.~\ref{Fig.6} shows that The student policy trained with TMP exhibits a lowest CoT. 
Specifically, across 4 different terrains, TMP achieves CoT reductions of 26.67\% and 2.384\% on discrete obstacles, 16.89\% and 2.205\% on rough slopes, 5.870\% and 14.35\% on slopes, 13.65\% and 6.604\% on stairs, compared to TS and ROA, respectively.
The student policies trained with TS and ROA exhibit a higher CoT, 
likely due to their reliance on supervised learning, which limits exploration capability. 
In contrast, TMP enables the student to dynamically learn the teacher’s strategy within the simulation environment.

\subsection{Real-World Experiments}
To evaluate the effectiveness and robustness of our control strategy, we conducted real-world experiments using the CASBOT SE humanoid robot. 
The experiments involved three distinct scenarios: traversing a sloped surface, walking over a rough brick-paved terrain, and responding to external disturbances. 
These tests demonstrate the robot’s adaptability to challenging environments and its ability to maintain stability under perturbations.

Fig.\ref{Fig.real} presents sequential snapshots of the experiments. 
In the initial sequence, the robot successfully traverses the sloped terrain by dynamically adjusting its joint configurations, particularly the foot pitch joints, to maintain balance. 
Taking this terrain as an example, the torque variations of the left leg are plotted on the right side of Fig.\ref{Fig.real}, where the six rows from top to bottom correspond to joints 1 to 6. 
It can be observed that during the transition from flat ground to a slope and back, the hip and knee joints exhibit relatively small torque variations, whereas the ankle pitch joint (second-to-last row) undergoes significant changes. 
This indicates that the proposed strategy ensures stable locomotion while achieving a lower Cost of Transport (CoT) and enhanced terrain adaptability. 
Moreover, the system effectively regulates joint torques in response to terrain inclination changes, mitigating undesired forward/backward tilting motions.

The second-row sequence shows the robot traversing a brick-paved terrain with discontinuous ground contact. 
Through adaptive foot placement and joint stiffness modulation, the robot compensates for terrain irregularities, maintaining upper body stability and dynamic balance despite unpredictable contact forces.

In the third row, the robot is subjected to external disturbances applied via sudden pushes. 
Upon receiving a perturbation, the robot swiftly reacts by adjusting its stepping strategy and redistributing its center of mass to regain balance. 
The control policy enables rapid recovery by leveraging proprioceptive feedback, ensuring stability even under sudden external forces.

These experiments validate the effectiveness of our approach in handling complex terrains and disturbances, highlighting the generalizability of our control strategy.

\section{Conclusions and Future Works}

The significance of this work lies in the novel framework design, which departs from the traditional teacher-student paradigm by eliminating the encoder structure and using a generative adversarial mechanism for knowledge transfer. 
It enables developers to easily train a teacher policy and transfer it to existing networks, improving performance without extensive restructuring. 
The framework also supports the future integration of exteroception modules, such as vision, without requiring retraining of the teacher policy.


\bibliographystyle{unsrt}  
\bibliography{Reference}  


\section*{APPENDIX}
\balance 
\begin{table}[H]
    \centering
    \caption{Structure of TMP Policy Networks. 
    The parts marked with an asterisk in the Aux network indicate the layers shared with the Actor.}
    \label{tab:teacher_student_policy}
    \begin{tabular}{c c}
        \toprule
            \textbf{Network} & \textbf{Structure} \\
        \midrule
            \multicolumn{2}{c}{\textbf{Teacher}} \\
        \midrule
            Actor & $[1440, 768, 512, 256, 128, 64]$   \\
            Critic & $[768, 256, 128]$   \\
        \midrule
            \multicolumn{2}{c}{\textbf{Student}} \\
        \midrule
            Actor & $[1440, 768, 256]$   \\
            Critic & $[768, 256, 128]$   \\
            Aux & $[1440^*, 768]$   \\
            Disc & $[256, 256, 128]$   \\
        \bottomrule
    \end{tabular}
\end{table}

\begin{table}[H]
    \centering
    \caption{Summary of Observation Space. 
    The table contains proprioception observations, privileged observations, and auxiliary observations. 
    The table also details their dimensions.}
    \label{tab:observations}
    \begin{tabular}{ c  c  c  c  c }
        \toprule
            \textbf{Component} & \textbf{Dims} & \textbf{Prop.} & \textbf{Priv.} & \textbf{Aux.} \\
        \midrule
            Clock Input & 2 & \checkmark & & \\
            Command & 3 & \checkmark & & \\
            Last Actions & 12 & \checkmark & & \\
            Joint Position & 12 & \checkmark & & \checkmark \\
            Joint Velocity & 12 & \checkmark & & \checkmark \\
            Base Angular Velocity & 3 & \checkmark & & \checkmark \\
            Euler Angles & 3 & \checkmark & & \checkmark \\
            Action Difference & 12 & & \checkmark & \checkmark \\
            Base Linear Velocity & 3 & & \checkmark & \checkmark \\
            Friction Coefficient & 1 & & \checkmark & \checkmark \\
            Contact Phase & 2 & & \checkmark & \checkmark \\
            Disturbance Force & 2 & & \checkmark & \\
            Disturbance Torque & 3 & & \checkmark & \\
            Gait Phase & 2 & & \checkmark & \\
            Body Weight & 1 & & \checkmark & \\
            Height Map & 187 & & \checkmark & \\
        \bottomrule
    \end{tabular}
\end{table}
\begin{table}[H]
    \centering
    \caption{Overview of Reward Function Composition. 
    This indicates the formula of the reward function and the corresponding weight coefficients.
    \textcolor{red}{The parts marked in red are used only during the student training phase.}}
    \label{tab:rewards}
    \begin{tabular}{ c c c }
    \toprule
    \textbf{Reward Term} & \textbf{Formula} & \textbf{Weight} \\
    \midrule
    Base Orientation & $\phi(P^b_{\alpha\beta}, 5)$ & 0.5 \\
    Default Joint Position & $\phi(\theta_t - \theta_0, 2)$ & 0.8 \\
    Base Height Tracking & $\phi(P^b_z - 1.1, 100)$ & 0.2 \\
    Velocity Mismatch & $\phi(\dot{P}^b_{z,\gamma,\beta} - \text{CMD}_{z,\gamma,\beta}, 5)$ & 0.5 \\
    Lin. Velocity Tracking & $\phi(\dot{P}^b_{xyz} - \text{CMD}_{xyz}, 5)$ & 1.4 \\
    Ang. Velocity Tracking & $\phi(\dot{P}^b_{\alpha\beta\gamma} - \text{CMD}_{\alpha\beta\gamma}, 5)$ & 1.1 \\
    Contact Forces & $\max(F_{L,R} - 400, 0, 100)$ & -0.05 \\
    Contact Pattern & $\phi(I_p(t) - I_d(t), \infty)$ & 1.4 \\
    Feet Clearance & $r_{fc}$ & 1.6 \\
    Collision & $n_{\text{collision}}$ & -0.5 \\
    Action Smoothness & $\|a_t - 2a_{t-1} + a_{t-2}\|_2$ & -0.003 \\
    Joint Acceleration & $\|\ddot{\theta}\|_2^2$ & -1e-9 \\
    Joint Torque & $\|\tau\|_2^2$ & -1e-9 \\
    Joint Power & $|\tau| \| \dot{\theta} \|$ & $2 \cdot 10^{-5}$ \\
    \textcolor{red}{Disc. Reward} & \textcolor{red}{$r_{disc}$} & \textcolor{red}{$2 \cdot 10^{-4}$} \\
    \bottomrule
    \end{tabular}
    \end{table}

\begin{table}[H]
    \caption{Overview of Domain Randomization. 
        Additive randomization increments the parameter by a value within the specified range while scaling
        randomization adjusts it by a multiplicative factor from the same range.}
    \centering
    \label{tab:domain_randomization}
    \begin{tabular}{ c  c  c  c }
    \toprule
    \textbf{Randomized Variable} & \textbf{Unit} & \textbf{Range} & \textbf{Operation} \\
    \midrule
    Friction & - & [0.2, 1.3] & Scaling \\
    Restitution & - & [0.0, 0.4] & Additive \\
    Push Interval & seconds & [8, $\infty$] & Scaling \\
    Push Velocity (XY) & m/s & [0, 0.4] & Additive \\
    Push Angular Velocity & rad/s & [0, 0.6] & Additive \\
    Base Mass & kg & [-4.0, 4.0] & Additive \\
    COM Displacement & meters & [-0.06, 0.06] & Additive \\
    Stiffness Multiplier & \% & [0.8, 1.2] & Scaling \\
    Damping Multiplier & \% & [0.8, 1.2] & Scaling \\
    Torque Multiplier & \% & [0.8, 1.2] & Scaling \\
    Link Mass Multiplier & \% & [0.8, 1.2] & Scaling \\
    Motor Offset & radians & [-0.035, 0.035] & Additive \\
    Joint Friction & - & [0.01, 1.15] & Scaling \\
    Joint Damping & - & [0.3, 1.5] & Scaling \\
    Joint Armature & - & [0.008, 0.06] & Scaling \\
    Lag Timesteps & steps & [5, 20] & Additive \\
    Observation Motor Lag & steps & [5, 20] & Additive \\
    Observation Actions Lag & steps & [2, 5] & Additive \\
    Observation IMU Lag & steps & [1, 10] & Additive \\
    Coulomb Friction & - & [0.1, 0.9] & Scaling \\
    Viscous Friction & - & [0.05, 0.1] & Scaling \\
    \bottomrule
    \end{tabular}
\end{table}

\begin{table}[H]
    \caption{Algorithm Environment Parameters. 
    \textcolor{red}{The parts marked in red are used only during the student training phase.}}
    \label{tab:environment_parameters}
    \centering
        \begin{tabular}{ c  c }
            \toprule
            \textbf{Environment Setting} & \textbf{Value}\\
            \midrule
            Observation Frames & 15 \\
            Privileged Observation Frames & 3 \\
            Number of Single Observation & 47 \\
            Number of Single Privileged Observation & 213 \\
            \textcolor{red}{Number of Single Auxiliary Observation} & \textcolor{red}{48} \\
            Height Measurement Range & 1.6m $\times$ 1m \\
            Number of Actions & 12 \\
            Number of Environments & 10240 \\
            Static Friction Coefficient & 0.6 \\
            Dynamic Friction Coefficient & 0.6 \\
            Terrain Block Size & 8m $\times$ 8m\\
            Terrain Levels & 20 \\
            Number of Terrains per Level & 20 \\
            \bottomrule
            \end{tabular}
    \end{table}

\begin{table}[H]
    \caption{Algorithm Framework Parameters.
    \textcolor{red}{The parts marked in red are used only during the student training phase.}}
    \label{tab:algorithm_parameters}
    \centering
        \begin{tabular}{ c  c }
            \toprule
            \textbf{Algorithm Parameter Setting} & \textbf{Value}\\
            \midrule
            Batch Size & 10240 $\times$ 24 \\
            Mini-batch Size & 10240 $\times$ 4 \\ 
            Value Function Loss Coefficient $\lambda_v$ & 1.0 \\
            Entropy Coefficient $\lambda_e$ & 0.001 \\
            \textcolor{red}{Disc. Loss Coefficient $\lambda_{\text{disc}}$} & \textcolor{red}{0.05}\\
            \textcolor{red}{Aux. Loss Coefficient $\lambda_{\text{aux}}$} & \textcolor{red}{0.1}\\
            \textcolor{red}{Prediction Loss Coefficient $\lambda_{\text{pred}}$} & \textcolor{red}{0.5}\\
            \textcolor{red}{Gradient Penalty Coefficient $\lambda_{\text{grad}}$} & \textcolor{red}{0.05}\\
            \textcolor{red}{Weight Decay Coefficient $\lambda_{\text{weight}}$} & \textcolor{red}{0.5} \\
            Learning Rate \(\alpha\) & 1e-3 \\
            Learning Rate Adjustment & adaptive / \textcolor{red}{fixed} \\
            Desired KL Divergence & 0.01 \\
            Clip Parameter & 0.1 \\
            Gradient Clipping Max Norm $\text{max\_grad}$ & 1.0 \\
            Learning Iterations per Epoch & 2 / \textcolor{red}{4} \\
            Discount Factor $\gamma$ & 0.995s \\
            GAE Discount Factor & 0.95 \\
            \bottomrule
        \end{tabular}
\end{table}

\addtolength{\textheight}{-6cm}   
\end{document}